\documentclass[10pt,twocolumn,letterpaper]{article}

\usepackage{multirow}
\usepackage{bbding}
\usepackage{textcomp}
\usepackage{siunitx}
\usepackage[pagenumbers]{cvpr} % To force page numbers, e.g. for an arXiv version

\definecolor{cvprblue}{rgb}{0.21,0.49,0.74}
\usepackage[pagebackref,breaklinks,colorlinks,allcolors=cvprblue]{hyperref}

\def\paperID{14789} % *** Enter the Paper ID here
\def\confName{CVPR}
\def\confYear{2025}

\title{LoyalDiffusion: A %Privacy-Preserving 
Diffusion Model Guarding Against Data Replication}

\author{Chenghao Li\\
\and
Yuke Zhang\\
\and
Dake Chen\\
\and
Jingqi Xu\\
\and
Peter A. Beerel\\
\and
University of Southern California\\
{\tt\small \{cli78217, yukezhan, dakechen, jingqixu, pabeerel\}@usc.edu} \\
}

\begin{document}
\maketitle
\begin{abstract}
Diffusion models have demonstrated significant potential in image generation. However, their ability to replicate training data presents a privacy risk, particularly when the training data includes confidential information. Existing mitigation strategies primarily focus on augmenting the training dataset, leaving the impact of diffusion model architecture under explored. In this paper, we address this gap by examining and mitigating the impact of the model structure, specifically the skip connections in the diffusion model’s U-Net model. We first present our observation on a trade-off in the skip connections. While they enhance image generation quality, they also reinforce the memorization of training data, increasing the risk of replication. To address this, we propose a replication-aware U-Net (RAU-Net) architecture that incorporates information transfer blocks into skip connections that are less essential for image quality. Recognizing the potential impact of RAU-Net on generation quality, we further investigate and identify specific timesteps during which the impact on memorization is most pronounced. By applying RAU-Net selectively at these critical timesteps, we couple our novel diffusion model with a targeted training and inference strategy, forming a framework we refer to as LoyalDiffusion. Extensive experiments demonstrate that LoyalDiffusion outperforms the state-of-the-art replication mitigation method achieving a $48.63\%$ reduction in replication while maintaining comparable image quality.

\end{abstract}    
\section{Introduction}
\label{sec:intro}

Diffusion models~\cite{ho2020denoising, dhariwal2021diffusion, rombach2022high} have emerged as powerful tools for generating high-quality images, with their performance further enhanced through conditional-guidance~\cite{ho2022classifier, rombach2022high}, a technique that allows models to generate images based on specific text input conditions, such as descriptive prompts or categorical labels. %like text or class labels. 
% This makes the models more controllable, producing tailored outputs in a wide range of applications. 
Starting with a pure noise image, prevalent diffusion models utilize a U-Net architecture to predict and remove the noise iteratively across a predetermined number of timesteps, gradually generating an image that corresponds to the specified input conditions~\cite{ho2022classifier, rombach2022high}.
Leading examples include Stable Diffusion~\cite{rombach2022high}, DALL·E~\cite{ramesh2021zero}, and Imagen~\cite{saharia2022photorealistic}, which have gained widespread recognition for their ability to generate art, photorealistic images, and illustrations from textual prompts~\cite{rombach2022high, huang2022draw}. 

%However, with the increasing use of diffusion models in sensitive fields such as healthcare, content generation, and proprietary image production, there are rising concerns about security and privacy~\cite{}. These concerns are especially relevant due to the potential for models to inadvertently leak sensitive information through various types of attacks. One critical privacy attack on diffusion models is training data extraction~\cite{carlini2023extracting, webster2023reproducible}, where models memorize specific samples from their training set and risk reproducing them during generation. Carlini et al.~\cite{carlini2023extracting} demonstrated this vulnerability by using a generate-and-filter pipeline to successfully extract training samples, highlighting that diffusion models may pose higher privacy risks compared to GANs. Additionally, membership inference attacks (MIA)~\cite{shokri2017membership} have emerged as a serious threat to diffusion models' privacy. Duan et al.~\cite{duan2023diffusion} introduced Step-wise Error Comparing Membership Inference (SecMI), which confirmed that diffusion models are susceptible to MIAs. Similar research efforts ~\cite{matsumoto2023membership, kong2023efficient} also illustrate this phenomenon.

%A significant factor in the vulnerability of diffusion models to privacy attacks is replication, which involves the model memorizing and regenerating specific training data. 

\begin{figure}
    \centering
    \includegraphics[width=\columnwidth]{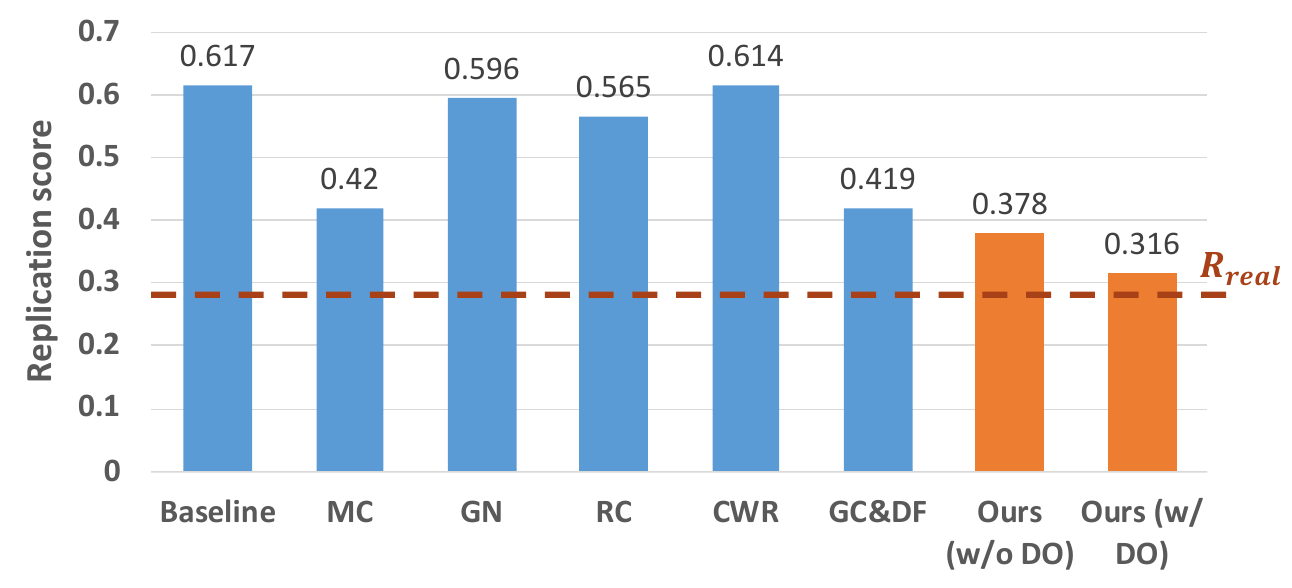}
    \caption{Comparison of replication scores between our proposed \textit{LoyalDiffusion} with and without dataset optimization (DO) and prior methods including MC~\cite{somepalli2023understanding}, GN~\cite{somepalli2023understanding}, RC~\cite{somepalli2023understanding}, CWR~\cite{somepalli2023understanding}, GC\&DF~\cite{li2024mitigate}.}
    \label{fig:comp_prior_work}
    \vspace{-5mm}
\end{figure}

However, a significant vulnerability of diffusion models is replication, where the model generates images that too closely resemble its training data~\cite{somepalli2023diffusion, somepalli2023understanding, carlini2023extracting}. 
This issue becomes particularly concerning when fine-tuning a large pre-trained model on %small
customized datasets~\cite{somepalli2023diffusion}, a common scenario in many applications, because %Small 
customized datasets often %dense with 
contain a substantial amount of
sensitive or private data which should not appear in the generated images~\cite{li2024shake}. 
%In this work, we focus on the replication issue during the fine-tuning phase.
% Somepalli et al.~\cite{somepalli2023diffusion, somepalli2023understanding} define 
%
%Replication is defined as the propensity of diffusion models to reproduce concepts, content, or styles embedded in their training data~\cite{somepalli2023diffusion, somepalli2023understanding}. 
%Replication arises because diffusion models inherently memorize patterns during fine-tuning, which enables high fidelity in generated images~\cite{}. However, such memory retention %comes at the risk of 
%also leads to the reproduction of exact visual elements from the fine-tuning set. %posing privacy threats when sensitive or proprietary information is involved. 
Previous studies~\cite{somepalli2023understanding, gu2023memorization, webster2023duplication, chen2024towards, li2024mitigate, wen2024detecting, dockhorn2022differentially, liu2023differentially, schramowski2023safe} have proposed various strategies to mitigate replication in diffusion models by optimizing the datasets. %, but each has notable limitations. 
Deduplication~\cite{somepalli2023understanding, webster2023duplication, chen2024towards, li2024mitigate}, for instance, helps reduce overfitting by removing duplicate content but is computationally demanding and struggles with large-scale datasets. Noise addition techniques~\cite{dockhorn2022differentially, liu2023differentially} including differential privacy~\cite{dwork2006differential} effectively limit privacy risks but introduce noise that degrades model quality. 
% Machine unlearning~\cite{}, which aims to erase specific content, can compromise model stability and result in degraded performance. 
Lastly, prompt disturbing~\cite{schramowski2023safe, wen2024detecting} modifies user inputs to prevent replication, yet this can lead to reduced control over output quality and introduce biases. These methods largely focus on external factors like datasets or inputs rather than the core model mechanisms that cause replication, highlighting the need for mitigating replication from the root, the model itself.

In this paper, we investigate replication in diffusion models from two new perspectives: the model structure and the timesteps during which the impact on memorization are significant but less important on generation quality. %training strategy, 
We propose a new diffusion model framework, referred to as \textit{LoyalDiffusion}, to mitigate the replication issue motivated by the hypothesis that 
%a method to mitigate replication while preserving model performance. Our approach is motivated by examining the U-Net architecture, the core of the diffusion model, which handles denoising iteratively across a large number of timesteps~\cite{}. Notably, 
the skip connections in the U-Net model may contribute to replication because they directly transfer the outputs of down-sampling blocks to up-sampling blocks. 
We first verify our hypothesis and discovered a trade-off with these skip connections: while they enhance the quality of image generation, they also reinforce the memorization of training data, thereby increasing the risk of replication. 
% Based on this hypothesis,
Thus motivated, we introduce a modified %de-replicate 
replication-aware U-Net structure, referred to as RAU-Net, that replaces the direct copying between down- and up-sampling layers with information transfer blocks.  %more complex operations. 
In particular, we propose several information transfer blocks and empirically select the one that minimizes replication while preserving the highest quality of image generation. Moreover, because replacing all skip connections with an information transfer block significantly degrades the quality of generated images, we empirically investigate the optimal placement for these blockers. 

Orthogonally, we show that the sensitivity to generation quality and replication varies across timesteps. Thus motivated, we 
%certain intervals have higher impact on FID and CLIP score and should not to be altered from the baseline U-Net to maintain high quality generation.
% certain intervals have minimal impact on replication and need not to be altered from the baseline U-Net.%and can be ignored. %Given that the de-replicate U-Net may reduce model performance, 
%Therefore, we 
selectively apply our RAU-Net to only a specific range of timesteps to optimize the balance between reduced replication and generation quality.
%and train both nets for their specific range of timesteps. 
Thus, LoyalDiffusion incorporates both the standard U-Net and our newly proposed RAU-Net that we apply (and train) for different ranges of timesteps. 

%Furthermore, we introduce a two-stage training strategy for optimizing our diffusion model. Particularly, different U-Nets are trained and used for inference upon different intervals of timesteps.
% multi-stage diffusion training strategy that minimizes the performance trade-offs associated with replication mitigation. 

Finally, we demonstrate that \textit{LoyalDiffusion} %de-replicate U-Net combined with multi-stage training 
can be effectively used alongside prior replication mitigation methods~\cite{li2024mitigate}, %dual-fusion enhancement~\cite{}, 
achieving near-optimal performance while substantially reducing replication risks. Figure~\ref{fig:comp_prior_work} compares our \textit{LoyalDiffusion} with prior methods showing that our approach achieves the smallest replication score.

%Not sure this paragraph
% Our study provides insights into the privacy-generation trade-offs inherent in model structure and training strategies, as illustrated in Figure \ref{}. The challenge of replication arises from the need to learn the underlying data distribution without overfitting to specific training samples. Ideally, a model should approximate the data distribution while avoiding memorization of individual samples, capturing general patterns rather than specific details. However, this goal is challenging; even cognitive systems, like the human brain, struggle to avoid memorizing specific data entirely. While complete mitigation of replication remains difficult, our approach focuses on reducing it as much as possible.

To the best of our knowledge, LoyalDiffusion is the first framework to consider addresses replication in diffusion models by  focusing on the U-Net structure and image generation evolvement, providing a novel perspective on replication mitigation. We summarize our contributions as follows:
% \begin{itemize}
% \setlength\itemsep{-0.5em}
(1) We propose a replication-aware U-Net (RAU-Net) which mitigates replication by limiting direct information flow between down-sampling and up-sampling. %This modification reduces the risk of memorization by disrupting straightforward feature copying, thereby encouraging more abstract feature learning while preserving data fidelity.
(2) We identify the impact of timesteps, and selectively apply the RAU-Net during those timesteps, minimizing replication while maintaining model performance. We also develop a corresponding training method to optimize model performance.
(3) Our approach outperforms prior methods by achieving the lowest replication score with minimal loss in image quality.
(4) We use Bing image search to obtain real world images that closely align with the tested prompts from which we estimate a lower bound of the replication score $R_{real}$ and find that our optimized models yield replication scores close to this estimate (see Figure \ref{fig:comp_prior_work}).

\section{Background and Related Works}

\begin{figure*}[t]
      \centering
      \includegraphics[width=\textwidth]{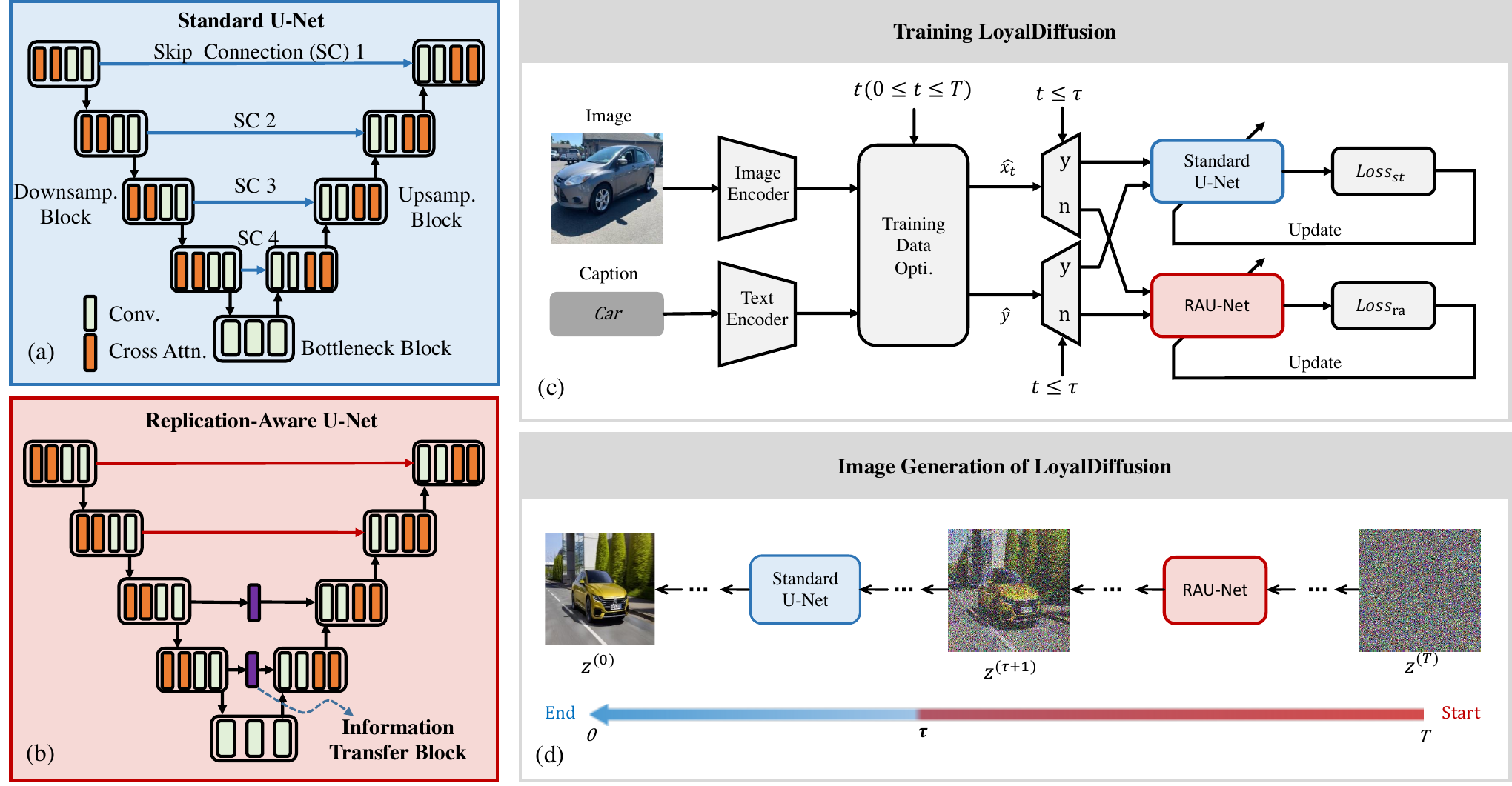}
      \caption{Overview of the proposed \textit{LoyalDiffusion} framework. (a) Standard U-Net architecture. (b) Replication-Aware U-Net (RAU-Net) where the skip connection (SC) 3 \& 4 being modified to one single Conv 3×3 layer. (c) Training strategy for \textit{LoyalDiffusion} where the training data optimization block includes generalized captioning and dual fusion~\cite{li2024mitigate}. (d) Image generation process using \textit{LoyalDiffusion}.}
      \label{fig:overview}
\end{figure*}

\subsection{Diffusion Models}

Denoising Diffusion Probabilistic Models (DDPMs)~\cite{ho2020denoising} have shown significant success in generative tasks~\cite{lugmayr2022repaint, ho2022video, lemercier2024diffusion}, which use a parameterized Markov chain to reverse a noise-adding process, reconstructing data from noise. However, their high computational complexity led to the development of Latent Diffusion Models (LDMs), such as Stable Diffusion~\cite{rombach2022high}, which reduce this cost by training in a compressed latent space. LDMs first encode images into a lower-dimensional latent space using an autoencoder, where a DDPM is then trained. Additionally, building on existing guidance techniques in diffusion models (DMs)~\cite{dhariwal2021diffusion, ho2022classifier}, LDMs employ cross-attention~\cite{vaswani2017attention} to enable multi-modal guidance by injecting conditional information at each timestep to better control generation.  

We refer to a pre-trained Stable Diffusion v2.1~\cite{rombach2022high} model as \emph{SD}, which is used as the baseline model for the rest of this paper. The %required training
fine-tuning dataset is typically formed with a number of image-caption pairs, denoted as $D_{FT}=\{(x^{n}, y^{n})\}_{n=1}^N$, where $x^{n}$ is the $n$-th image, and $y^{n}$ is its corresponding caption. 
Training a diffusion model involves two phases: forward process and reverse training process. 
The forward process adds noise iteratively to each training image $x$ over $T$ timesteps, gradually transforming it into pure Gaussian noise. %$x_{T}$. 
We use $x_{t}$ and $\epsilon_t$ to denote the noisy input data and the noise added to the input at time step $t$, respectively.
In the reverse training process, \emph{SD} learns to reverse this transformation, gradually denoising the noisy data to reconstruct the original input. Specifically,
%In the reverse process, we aim to gradually denoise $x_{T}$ back to $x_{0}$. 
a U-Net parameterized by $\theta$ learns to estimate the noise $\epsilon_t$ %\epsilon_{\theta}(x_{t},t)$ 
added at timestep $t$, and \emph{SD} can then subtract the predicted noise from $x_t$ to iteratively refine the data. The loss function for training the U-Net is as follows:
% \begin{equation}
%     L(\theta) = E_{\varepsilon, t}[\|\varepsilon - \varepsilon_{\theta}(x_{n}^{(t)}, t, y_{n})\|^2].
% \end{equation}
\begin{equation}
    L(\theta) = E_{\varepsilon, t}[\|\epsilon_t - \varepsilon_{\theta}(x_t, t, y)\|^2],
\end{equation}
where $\varepsilon_{\theta}(x_t, t, y)$ represents the noise predicted by U-Net for the noisy data $x_t$ at time step $t$ conditioned on the caption $y$.

\subsection{U-Net and Skip Connection}

As shown in Figure \ref{fig:overview}(a), U-Net~\cite{ronneberger2015u} is a symmetric, encoder-decoder architecture with skip connections, which play a crucial role in diffusion models. The encoder path consists of multiple convolutional layers and cross-attention blocks that perform downsampling operations, progressively reducing the spatial resolution while capturing increasingly abstract features. 
At the center of the network, the bottleneck provides a highly compressed, low-level representation of the input data, containing the most essential features.
The decoder path then uses upsampling layers to progressively reconstruct the spatial dimensions back to the original resolution. In this upsampling path, each layer’s output is concatenated with the corresponding feature map from the encoder via skip connections. 

Skip connection were introduced to mitigate 
the degradation problem by helping stabilize gradients during training and enhancing information flow across layers~\cite{he2016deep, xu2024development}, and have been applied to various U-Net designs~\cite{ronneberger2015u, badrinarayanan2017segnet, zhou2019unet++, mao2016image}.  
These skip connections help retain high-resolution details from the encoder, enabling more precise reconstructions and improving the model's ability to denoise, segment or reconstruct the data accurately.
However, while skip connections improve feature retention, they may also increase the risk of replication by enabling the diffusion model to memorize specific details from training data~\cite{wen2024detecting}.

% \subsection{Replication Score and Mitigation Strategies}
\subsection{Replication and Mitigation Strategies}
% Despite their success, diffusion models are vulnerable to unintended memorization and replication of training data~\cite{somepalli2023diffusion, somepalli2023understanding, li2024mitigate, carlini2023extracting, duan2023diffusion}, which raises concerns about privacy and data security in sensitive applications.
% The generative process is expressed as $g_{n} = SD(p_{n}, r)$, where $g_{n}$ represents the generated image, $p_{n}$ represents the corresponding caption, and $r$ represents the initial random noise. Building upon Somepalli et al.'s research~\cite{somepalli2023understanding}, $X=\{x_{n}\}_{n=1}^N$ and $G=\{g_{n}\}_{n=1}^N$ exist significant object-level copying, defined as replication. 

We use $X=\{x^{n}\}_{n=1}^N$ and $G=\{g^{n}\}_{n=1}^N$ to represent the training image set and the generated image set, respectively. The degree of replication between $X$ and $G$ can be evaluated using the replication score~\cite{somepalli2023understanding}, which quantifies the extent to which generated images replicate or closely resemble images from the training set.

\textbf{Replication Score.} %How to quantify data replication is complicated as it may refer to a diverse set of notions including object-level, semantic-level, and style-level replication. To simplify the evaluation process, we follow~\cite{somepalli2023understanding, somepalli2023diffusion} which focuses on object-level replication. In particular, in order for comparative analyses with existing research, we evaluate replication based on the metrics proposed by~\cite{somepalli2023understanding}. 
Let $S$ represent the set of similarity scores of all possible pairs of images $x\in X$ and $g \in G$. The replication score, denoted as $R$, is defined as the $95$th percentile of the similarity score distribution. Formally,
%threshold satisfying
% By calculating the feature map similarity between dataset $X$ and $G$, the probability density function (PDF) of $S$ can be obtained as $PDF_{S}(S)$ and its cumulative distribution function (CDF) is $cdf_{S}(S<s)$, where random variable $S$ represents the similarity score. Since too many low-similarity images may distort the similarity of the entire dataset, only the top $5\%$ of the similarity distribution is taken into account. This threshold $R(X, G)\in{[0, 1]}$ is represented as the dataset replication score, defined as:
\begin{equation}
     % Pr(S < R) = cdf_{S}(S < R) = 0.95.
    Pr(s < R) = 0.95, s\in S.
\end{equation}
This means that $95\%$ of the similarity scores in $s \in S$ are below the replication score $R$, ensuring that low similarity scores do not distort the overall dataset replication score. The similarity score between two images is calculated based on features extracted by the SSCD~\cite{pizzi2022self}. 
A higher value of $R$ indicates a more pronounced degree of replication.
% Somepalli et al.~\cite{somepalli2023diffusion, somepalli2023understanding} introduce and define the concept of replication in diffusion models as the memorization of specific details from training images. Replication is largely driven by image duplication in the training data~\cite{somepalli2023understanding}. Furthermore, highly specific captions act as keys that retrieve specific data points from the model’s memory~\cite{somepalli2023understanding, li2024mitigate}. This issue is further intensified by smaller datasets, where limited diversity increases the likelihood of memorization~\cite{somepalli2023understanding}. 

\textbf{Mitigation Methods.} As prior research indicates that replication is largely driven by image duplication~\cite{somepalli2023understanding} and highly specific captions~\cite{somepalli2023understanding, li2024mitigate}  in the training data, %. Furthermore, highly specific captions act as keys that retrieve specific data points from the model’s memory~\cite{somepalli2023understanding, li2024mitigate}. 
existing mitigation strategies primarily focus on optimizing the training data. \cite{webster2023duplication, abbas2023semdedup} apply image deduplication to identify and remove both exact duplicates and semantically similar images. To generalize captions, Somepalli et al. propose methods including using multiple captions to training DMs (MC), adding gaussian noise to text embeddings (GN), randomly replacing captions (RC), randomly replacing or adding tokens or words in captions (RTA), repeating a random word in random location (CWR), and randomly adding a number to captions (RNA)~\cite{somepalli2023understanding}. Additionally, Li et al. introduce a generality score to quantify caption abstraction and use LLM~\cite{achiam2023gpt} to generalize captions, along with a dual-fusion technique that merges each training image with an external image-caption pair based on a set weighting, thereby reducing memorization risks. This approach focuses on a combination of caption generalization and image deduplication, thereby significantly reduces replication. Anti-Memorization Guidance (AMG)~\cite{chen2024towards} includes three guidance strategies, despecification guidance, caption deduplication guidance, and dissimilarity guidance, all aiming at addressing memorization arising from image and caption duplication and over-specification. Wen et al.~\cite{wen2024detecting} suggest using perturbed prompt embeddings to minimize text-conditional prediction magnitude at diffusion inference, and screen out potential memorized pairs based on prediction magnitude during training. However, these data-centric strategies predominantly address replication externally, without examining the model’s internal mechanisms that contribute to memorization. %Moreover, differential privacy~\cite{dockhorn2022differentially, liu2023differentially} has also been applied to DMs, but it can degrade the model's output quality and reduce fidelity in generated images. 
Machine unlearning~\cite{bourtoule2021machine} is able to erase and remove specific content from the models and several studies have introduced it to mitigate replication~\cite{zhang2024forget, kumari2023ablating}. However, it can be computationally expensive and has potential to introduce new vulnerabilities. 

To address those shortcomings, in this paper, we propose a novel approach to mitigate replication by focusing on the diffusion model structure, specifically the U-Net architecture and training strategy.

\section{The Proposed LoyalDiffusion Framework}

Our approach to reducing replication in diffusion models, as shown in Figure \ref{fig:overview}, 
has three components. 
First, we introduce a replication-aware U-Net (RAU-Net), which replaces the 
U-Net’s skip connections with more complex
%convolutional layers
information transfer block, mitigating the transference of direct features between the encoder and decoder of the model. Secondly, we propose a two-stage training strategy that applies the RAU-Net only at higher timesteps, 
% where we hypothesize replication risks are greatest, 
where the impact on fidelity and quality of generation is minimal while still maintaining significant replication reduction,
using a standard U-Net at lower timesteps to avoid fidelity of finer details being destroyed. 
Finally, we show how these techniques can be combined with generalized captioning and dual-fusion enhancement to further improve generalization and sample quality.

\subsection{Replication-Aware U-Net}
\label{sec:3-1}

% In diffusion models that use U-Net architectures, skip connections play a crucial role in preserving spatial information by connecting corresponding layers of the encoder and decoder. Typically, these skip connections are identity functions that pass encoder features directly to the decoder.  We hypothesize that these direct transfer can lead to replication in the generated images. 
\textbf{Skip Connection (SC) and Replication.} 
Skip connections in U-Net architectures are designed to preserve spatial features from the encoder by passing them directly to the decoder, effectively bypassing the bottleneck layer~\cite{wang2022residual, anwar2019real, xu2024development}. This direct transfer helps retain high-resolution details that might otherwise be lost during the encoding and decoding processes. However, the bypassing also means the decoder block relies less on the most compressed and abstract features from the bottleneck layer which help reduce the reliance on specific features. 
% By allowing unaltered feature maps to flow directly to the decoder, the network risks that these detailed inputs may dominate the abstracted, high-level representations.
As a result, the decoder can become overly dependent on these unaltered encoder features. Instead of learning to generate new content based on latent representations, the decoder might start copying details directly from the encoder—a "copy-paste" behavior. This mechanism impacts the information flow by allowing high-resolution details to take precedence, limiting the network's ability to learn general, abstract representations, 
%This direct copying reduces the diversity and novelty of the generated content, 
making the model more prone to reproducing specific features from the training data rather than creating new, varied outputs.

To test this hypothesis, we compare the replication score of $SD$ with and without skip connections, as shown in Table~\ref{tab:raunet-skipconnection}. We use the Frechet Inception Distance (FID)~\cite{lucic2018gans} to assess the quality and diversity of generated images. The detailed experimental setup is provided in Section~\ref{sec:experiment_setup}. Based on the results, we conclude that removing skip connections can indeed reduce the replication score, however, at the expense of generated image quality. 

\begin{table}[]
\centering
\resizebox{0.5\columnwidth}{!}{%
\begin{tabular}{c|cc}
\hline\hline
     & Baseline & w/o SC      \\ \hline
R$\downarrow$   & 0.617  & 0.232   \\ \hline
FID$\downarrow$   & 18.01   & 127.73  \\ \hline \hline
% CLIP$\uparrow$ & -       & 0.313        & -       & -       & -       & -         \\ \hline
\end{tabular}%
}
\caption{Replication score and FID for the baseline diffusion model with and without skip connections on LAION-10K~\cite{schuhmann2022laion}.}
\label{tab:raunet-skipconnection}
\end{table}

\textbf{Information Transfer Block.}
% To mitigate the replication caused by direct skip connections in U-Net architectures, 
To maintain image generation quality, we propose adding information transfer blocks to selective skip connections rather than removing all skip connections indiscriminately. The information transfer block candidates include:

\begin{enumerate}
    % \item Removing Skip Connections (RSK): Eliminating skip connections entirely forces the decoder to rely only on the bottleneck's compressed representations, reducing replication risk. However, this may also lose important spatial information, potentially affecting image detail.
    \item Max Pooling (\textit{MP}): Max pooling reduces the spatial dimensions of features in the skip connections, emphasizing prominent elements and condensing encoder information. It helps prevent the decoder from accessing detailed patterns, while still preserving essential spatial information. Specifically, we use Max pooling with kernel size $2\times2$, followed by a \textit{MaxUnpool} that set all no-maximal values to $0$.
    
    \item Convolutional Layer (\textit{Conv}): Adding a convolutional layer to the skip connections allows the model to transform encoder features before passing them to the decoder, suppressing redundant patterns that could lead to replication. By adjusting the features, the convolutional layer helps balance the detailed information and the abstract representations. For \textit{Conv}, we use a single convolutional layer with kernel size $3\times3$, without changing the output feature shape.
    
    \item Multiple Convolutional Layers (\textit{Multi-Conv}): Extending \textit{Conv}  with multiple convolutional layers offers deeper feature processing, creating complex abstractions to further reduce replication artifacts. %However, this approach may increase computational costs and degrade image quality.
    We use $3$ Convolutional layers with kernel size $3\time3$.%, serving as downsampling, bootleneck, and upsampling for Multi-Conv. 
\end{enumerate}

% We analyze the effectiveness of different information transfer blocks for reducing replication. In this evaluation, we applied each block uniformly to all skip connections within the U-Net. The results are shown in Table~\ref{tab:raunet-result-1}.

We analyze the effectiveness of different information transfer blocks for reducing replication. %In this evaluation, we applied each block uniformly to all skip connections within the UNet. The results are shown in Table~\ref{tab:raunet-result-1}.
For this evaluation, we applied each type of block uniformly across all skip connections within the U-Net and also tested each type on a single skip connection. The results are presented in Table~\ref{tab:raunet-result-2}. The results indicate that the \textit{Conv} block provides the most effective mitigation against replication compared to other alternatives, leading us to select it as the information transfer block in our RAU-Net. 

Noting that applying the information transfer block to all skip connections results in either high FID or minimal replication improvement,  we selectively apply the \textit{Conv} block to specific skip connections in RAU-Net. As shown in Table~\ref{tab:raunet-result-2}, applying \textit{Conv} to SC $3$ achieves the lowest replication score while maintaining a comparable FID, leading us to adopt this design in RAU-Net. Interestingly, modifying SC $4$ using a \textit{Conv} resulted in almost no change in both replication and FID. This suggests that the information carried by SC 4 may not contribute significantly to replication issues. Given this finding, we experimented with modifying both SC 3 and SC 4 together. The result showed a replication score of $0.386$, further improving upon modifying only SC 3 by $13.07\%$ and $37.44\%$ better than baseline, without increasing FID. We thus apply \textit{Conv} to both SC $3$ and SC $4$ in RAU-Net as shown in Figuer~\ref{fig:overview}(b). We present experimental results on the impact of applying the information transfer block to various skip connection combinations in the Appendix.

% \begin{table*}[]
% \centering
% \resizebox{\textwidth}{!}{%
% \begin{tabular}{c|ccccc|ccccc|ccccc|ccccc}
% \hline
%         & \multicolumn{5}{c|}{RSK}             & \multicolumn{5}{c|}{MP}           & \multicolumn{5}{c|}{Conv}                      & \multicolumn{5}{c}{Multi-Conv}     \\ \hline
% Skip group         & 1       & 2      & 3       & 4  & All      & 1      & 2      & 3      & 4  & All     & 1       & 2       & 3       & 4       & All   & 1      & 2      & 3       & 4   & All    \\ \hline
% R$\downarrow$           & 0.570 & 0.583 & 0.601 & 0.621 & 0.232& 0.614 & 0.608 & 0.602 & 0.617 & 0.581 & 0.450 & 0.561 & 0.444 & 0.628 &0.168 & 0.537 & 0.582 & 0.511 & 0.620 &0.194 \\ \hline
% FID$\downarrow$           & 22.96  & 22.60 & 21.19  & 19.67 & 127.73  & 19.77 & 19.17 & 17.59 & 18.65 & 19.65 & 44.58  & 22.36  & 21.37  & 18.00   &  200.37 & 27.34 & 20.96 & 21.11  & 18.29 & 235.50 \\ \hline

% % CLIP$\uparrow$        & 0.313        & 0.307       & 0.304      & 0.306       & 0.313     & 0.313      & 0.314      & 0.314      & 0.314        & 0.287        & 0.301       & 0.290       & 0.314       &    0.304   & 0.302      & 0.306      & 0.298       & 0.313      \\ \hline
% \end{tabular}%
% }
% \caption{}
% \label{tab:raunet-result-2}
% \end{table*}

\begin{table}[]
\centering
\resizebox{0.95\columnwidth}{!}{
\begin{tabular}{l|cccccr}
\hline  \hline
    \multicolumn{6}{c}{Removing Skip Connection}     \\ \hline
% RSK             &       &       &       &       &       &       \\
SC \#      & 1     & 2     & 3     & 4     & All   &       \\\hline
R$\downarrow$               & 0.570 & 0.583 & 0.601 & 0.621 & 0.232 &       \\\hline
FID$\downarrow$            & 22.96 & 22.60 & 21.19 & 19.67 & 127.73 &       \\ \hline \hline
 \multicolumn{7}{c}{\textit{MP}}   \\\hline
SC \#      & 1     & 2     & 3     & 4     & All   &       \\ \hline
R$\downarrow$              & 0.614 & 0.608 & 0.602 & 0.617 & 0.581 &       \\ \hline
FID$\downarrow$              & 19.77 & 19.17 & 17.59 & 18.65 & 19.65 &       \\ \hline  \hline
 \multicolumn{7}{c}{\textit{Conv}}  \\ \hline
SC \#      & 1     & 2     & 3     & 4     & All   &       \\ \hline
R$\downarrow$             & 0.450 & 0.561 & \textbf{0.444} & 0.628 & 0.168 &       \\ \hline
FID$\downarrow$              & 44.58 & 22.36 & \textbf{21.37} & 18.00 & 200.37 &       \\ \hline  \hline
 \multicolumn{7}{c}{\textit{Multi-Conv}}      \\ \hline
SC \#      & 1     & 2     & 3     & 4     & All   &       \\ \hline
R$\downarrow$             & 0.537 & 0.582 & 0.511 & 0.620 & 0.194 &       \\ \hline
FID$\downarrow$              & 27.34 & 20.96 & 21.11 & 18.29 & 235.50 &       \\ \hline  \hline
\end{tabular}}
\caption{Replicaiton score and FID for different information transfer blocks on LAION-10K~\cite{schuhmann2022laion}. The 'SC \#' indicates the information transfer block is only applied to the specific skip connection numbered as in Figure~\ref{fig:overview}(a)}
\label{tab:raunet-result-2}
\end{table}

\subsection{Training \textit{\textbf{LoyalDiffusion}}}

\begin{figure}[t]
      \centering
      \includegraphics[width=\columnwidth]{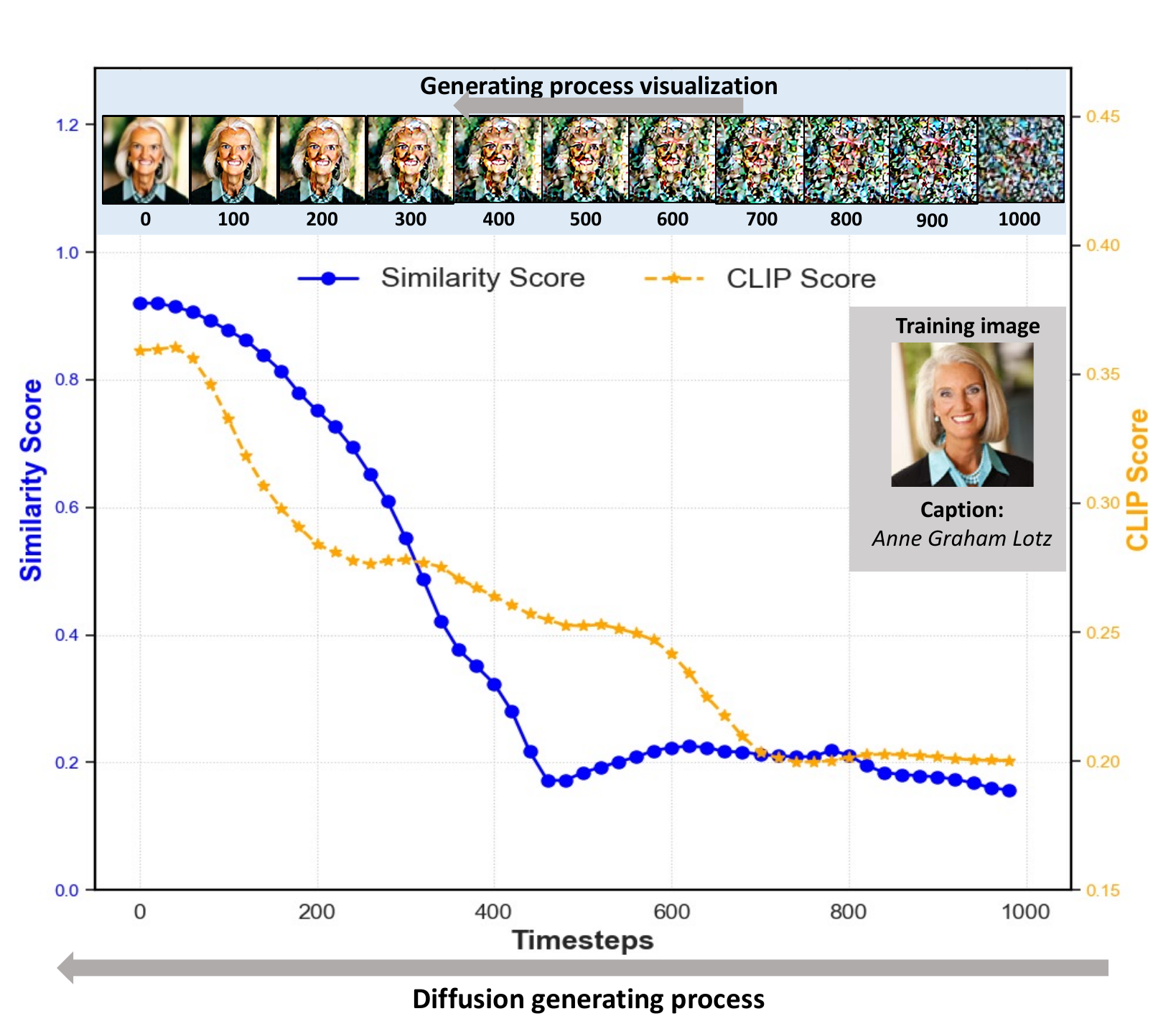}
      \caption{Similarity between a training image and the generated image that replicates it and CLIP score calculating using the caption during the diffusion generating process.  It shows that similarity and CLIP score do not change too much during the early inference interval with large timesteps.}
      \label{fig:two-stage-approach}
\end{figure}

%Building upon our efforts to mitigate replication artifacts through entropy reduction in skip connections, 
We recognize that applying the RAU-Net uniformly across all timesteps in the diffusion process may negatively impact overall model performance. To address this, we propose a two-stage diffusion training approach that strategically applies RAU-Net only at 
% the most influential timesteps for replication or 
the least influential timesteps for generation quality, thereby balancing the trade-off between replication reduction and model fidelity.

As shown in Figure \ref{fig:two-stage-approach}(c), we analysis the behavior of generating process for a highly replicated generated image. We use pairwise similarity score to represent the quality of generated image and CLIP score~\cite{radford2021learning} to manage how well it matches the prompt.  The similarity increases most rapidly during the later timesteps, when $t$ is small, but is stable at early timesteps, which suggests that early timesteps are less important to image quality than later timesteps. The CLIP score curve shows a similar trend, remaining almost unchanged during early timesteps.
Intuitively, during these early timesteps, the noise level remains high, making it difficult for the model to capture detailed features that define the generated image quality or precise alignment with the prompt. At this stage, the model primarily operates in a generalized manner, with limited specificity, as the high noise overwhelms finer details from the training data. This observation suggests that these early timesteps are less critical for maintaining high-quality generation results. By strategically applying RAU-Net only during these high-noise, low-impact timesteps, we can minimize any potential adverse effects on generating quality while still benefiting from RAU-Net's replication reduction capabilities. 

On the other hand, during the early timesteps, the model begins by generating high-level information, such as object outlines and shapes, as shown in Figure \ref{fig:two-stage-approach}. These foundational elements establish the general direction of the subsequent generation process. Given that our study's replication score is based on object-level, we hypothesize that this high-level information generated in the early stages is a critical factor in replication.

% the model generates high-level information from complete noise, such as object outlines and shapes, which establish the general direction of the subsequent generation process. In contrast, the later timesteps refine detailed information. Since the replication score used in our study is based at the object-level, we hypothesize that the high-level information generated in the early stages plays a crucial role in replication, whereas subsequent detailed refinements have a limited impact.

% As shown in Figure \ref{fig:two-stage-approach}, the similarity between generated samples increases most rapidly during the initial timesteps, suggesting that replication primarily occurs at larger timesteps. Intuitively, during these early timesteps, when $t$ is large, the model generates high-level information from complete noise, such as object outlines and shapes, which establish the general direction of the subsequent generation process. In contrast, the later timesteps refine detailed information. Since the replication score used in our study is based at the object-level, we hypothesize that the high-level information generated in the early stages plays a crucial role in replication, whereas subsequent detailed refinements have a limited impact.

Based on these insights, we propose a two-stage training strategy with a predefined threshold timestep $\tau$. When $t > \tau$, the RAU-Net $U_{ra}$ is employed and only trained on noisy training samples within $t\in(\tau, T)$. For $t \leq \tau $, the standard U-Net $U_{st}$ is used and only trained on $t\in[0, \tau]$ training samples. 

Mathematically, the denoising process at each timestep $t$ can be defined as:
\begin{equation}
x_{t-1} = \begin{cases}
U_{\text{ra}}(x_t, t), & \text{if } t > \tau, \\
U_{\text{st}}(x_t, t), & \text{if } t \leq \tau,
\end{cases}
\end{equation}
where $x_{t}$ is the noisy image at timestep $t$.
During training, both U-Nets are optimized concurrently but each is responsible for different ranges of timesteps. 
% Specifically, $U_{\text{standard}}$ is trained on noisy images from $x_{0}$ to $x_{\tau}$, while $U_{\text{de-rep}}$ is trained on noisy images from $x_{\tau+1}$ to $x_{T}$. 

% The total loss is the sum of the losses from both U-Nets:
% \begin{equation}
% L(\theta_{\text{st}}, \theta_{\text{ra}}) = L(\theta_{\text{st}}) + L(\theta_{\text{ra}})
% \end{equation}

%In summary, this two-stage approach allows us to selectively apply the de-replicate U-Net during the most critical timesteps, thereby minimizing adverse effects on overall model performance.

The image generation procedure of \textit{LoyalDiffusion} is illustrated in Figure~\ref{fig:overview}(d), where RAU-Net is utilized for timesteps greater than the threshold, while the standard U-Net is applied for timesteps below the threshold.

\subsection{Layering Dataset Optimization Methods}

To further enhance replication mitigation, we show our approach can be integrated with the state-of-the-art dataset optimization methods, i.e., generalized caption and dual-fusion techniques (GC\&DF)~\cite{li2024mitigate}. Generalized captioning reduces replication by transforming specific captions into more generalized descriptions using a large language model, thereby encouraging the model to learn broad patterns instead of memorizing specific details. Dual-fusion further supports this by fusing a training image with another image from a different resource, using a weighted fusion of both the images and their captions. These approaches tend to diversify the training data, making it harder for the model to replicate exact training samples. Together with our LoyalDiffusion, this combination addresses replication from both data and model perspectives.

\subsection{Evaluation using Bing Search Images}
\label{sec:RepliBing}

To further evaluate replication, we utilize the Bing search engine to retrieve images from the internet using the same prompts used during training and inference, forming a new dataset named \textit{RepliBing}, for which we calculate a replication score $R_{real}$ comparing \textit{RepliBing} to our training images. Because the \textit{RepliBing} dataset is diverse, closely aligns with the provided prompts, and is unlikely to have direct replications of training images 
(an assumption we verify manually), we assert $R_{real}$ is 
a good estimate of the lower bound of the replication score. 
In fact, as described in Section \ref{sec:experiment}, we 
found the replication score of all generative models tested is above $R_{real}$, but for our proposed models not by much, suggesting our models avoid replication well.

%This approach helps estimate a lower bound for replication scores.  We aim to mitigate replication by ensuring that the generated images resemble real world images, as real world images are unlikely to be direct replications of one another. 

% \input{sec/6_experiments}
\section{Experiments}\label{sec:experiment}

\subsection{Setup}\label{sec:experiment_setup}
To evaluate our strategy, we fine-tune large pre-trained models on small datasets, as this approach tends to intensify data replication issues~\cite{somepalli2023understanding}. This worst-case scenario allows us to assess the limits of our approach in mitigating replication.

\textbf{Baseline.} We use Stable Diffusion v2.1~\cite{rombach2022high} as our baseline model, which is currently one of the state-of-the-art text-to-image generation models trained on the LAION dataset~\cite{schuhmann2022laion}. Stable Diffusion consists of: a text encoder, an image autoencoder, a diffusion module, and an image decoder. For our experiments, we set the timestep with $T=1000$, focusing specifically on fine-tuning U-Net.

\textbf{Dataset.} For training, we randomly sample 10,000 instances from the LAION-2B~\cite{schuhmann2022laion} dataset. Each instance contains an image paired with its corresponding caption, providing a diverse set of content covering various objects and contexts.

\textbf{Training and Evaluation.} In our setup, all components except the U-Net are frozen during fine-tuning. We adopt most of the hyperparameters from~\cite{somepalli2023diffusion, somepalli2023understanding}, including training for 100,000 iterations with a learning rate of $5e^{-6}$. For evaluation, we generate 10,000 images and use three primary metrics: the replication score R to quantify the extent of replication in generated samples, the Frechet Inception Distance (FID)~\cite{lucic2018gans} to assess the quality and diversity of generated images, and the CLIP score~\cite{radford2021learning} to evaluate the semantic similarity between generated images and their corresponding captions.

\textbf{Real World Replication Score.} We calculate the replication score between \textit{RepliBing} and the training image set, resulting in a replication score of $R_{real}=0.275$. Interestingly, the replication score is not zero because even real-world images may share similar high-level structures or features, especially when generated from similar prompts.

\begin{table*}[]
\centering
\resizebox{\textwidth}{!}{%
\begin{tabular}{c|ccccccccccc}
\hline\hline
 % & \begin{tabular}[c]{@{}c@{}}Base-\\ line\end{tabular} & MC~\cite{somepalli2023understanding}    & GN~\cite{somepalli2023understanding}    & RC~\cite{somepalli2023understanding}    & CWR~\cite{somepalli2023understanding}   & GC\&DF~\cite{li2024mitigate} & \begin{tabular}[c]{@{}c@{}}Ours w/o\\ GC\&DF\end{tabular} & \begin{tabular}[c]{@{}c@{}}Ours w.\\ GC\&DF\end{tabular} \\ \hline
 &$R_{real}$ & Baseline & MC~\cite{somepalli2023understanding}& GN~\cite{somepalli2023understanding} & RC~\cite{somepalli2023understanding}& CWR~\cite{somepalli2023understanding} & GC\&DF~\cite{li2024mitigate} & Ours w/o GC\&DF& Ours w/ GC\&DF \\ \hline
R$\downarrow$ & 0.275       & 0.617                                                & 0.420 & 0.596 & 0.565 & 0.614 & 0.419  & 0.378                                                     & \textbf{0.316}                                                    \\ \hline
FID$\downarrow$ & -      & 18.01                                                & 16.83 & 19.50 & 15.98 & 16.73 & 17.16 & 19.17                                                     & 18.98                                                    \\ \hline \hline
\end{tabular}%
}
\caption{Compare \textit{LoyalDiffusion} with prior works.}
\label{tab:compare-prior}
\end{table*}

\subsection{Comparison with Prior-Art}

We compare our approach with a two-stage training threshold of $\tau=300$ with several prior-art methods designed to mitigate replication, including multiple captions (MC)~\cite{somepalli2023understanding}, Gaussian noise (GN)~\cite{somepalli2023understanding}, random caption replacement (RC)~\cite{somepalli2023understanding}, caption word repetition (CWR)~\cite{somepalli2023understanding},
% , random token addition (RTA)
 Generalized Captions and Dual-Fusion Enhancement (GC\&DF)~\cite{li2024mitigate}.
% , and other state-of-the-art methods (placeholder for future methods). 
The results are shown in Table \ref{tab:compare-prior}. 
% Each method aims to address replication from different perspectives, either by modifying the input data or by changing the model architecture or training strategy. 
Specifically, our method outperforms the best prior-art method, GC\&DF, by $24.58\%$, without significantly impacting FID.
Moreover, our best-performing model achieves a replication score of 0.316 that is close to the real world replication score $R_{real}$ of 0.275, indicating that our approach is highly effective in mitigating replication.

\subsection{Two-Stage Training Threshold Analysis}

Building on the effective reduction of replication achieved by RAU-Net, which replaces skip connection groups 3 and 4 with \textit{Conv}, we selected this architecture for further experimentation.
% To evaluate a multi-stage training strategy, we employ both the standard U-Net $U_{st}$ and the RAU-Net $U_{ra}$ and a threshold timestep $\tau$.
%
To evaluate our two-stage training strategy, the standard U-Net $U_{st}$ is trained on timesteps $t \leq \tau$, while the RAU-Net $U_{ra}$ handles timesteps $t > \tau$.
% the alternative setup reverses this arrangement. This design aims to determine whether larger or smaller timesteps have a greater impact on replication. By understanding which stages each architecture performs best, we can improve the overall effectiveness of the denoising process. 
Then we test different $\tau\in\{100, 300, 500, 700, 900\}$ and show the results in Table \ref{tab:two-stage-result_1}.
% and \ref{tab:two-stage-result_2}.

\begin{table}[]
\centering
\resizebox{0.95\columnwidth}{!}{%
\begin{tabular}{c|ccccc}
\hline\hline
$\tau$ & 100    & 300             & 500    & 700    & 900    \\ \hline
R$\downarrow$                & 0.373 & \textbf{0.378} & 0.453 & 0.558 & 0.606 \\ \hline
FID$\downarrow$              & 22.08 & \textbf{19.17} & 18.89 & 18.33 & 16.77 \\ \hline
CLIP$\uparrow$             & 0.290      & \textbf{0.301}               & 0.306      & 0.313      & 0.316      \\ \hline\hline
\end{tabular}%
}
\caption{Results for switching different $\tau$, with $U_{st}$ for $t\leq \tau$ and $U_{ra}$ (\textit{Conv} on skip connection group 3 \& 4) for $t>\tau$ without GC\&DF.}
\label{tab:two-stage-result_1}
\end{table}

% \textbf{Analysis} 
We see that the best performance is achieved when 
$\tau=300$. This configuration results in an $38.74\%$ reduction in the replication score compared to the baseline model and an $2.07\%$ reduction compared to training without two-stage strategy. Additionally, this approach reduces the FID score by $2.14$ compared to using only the $U_{ra}$ (that yields a FID of $21.31$ as seen in Table~\ref{tab:ablation-1}, indicating improved model generation ability.

% We also plotted the replication score and FID versus timesteps during inference to understand the impact on different stages, as shown in Figure x. These plots, comparing the baseline model, the de-replicate U-Net, and the best-performing multi-stage model, show that ...........

\subsection{Impact of Layering Data Set Optimization}

\begin{figure}[tbhp]
      \centering
      \includegraphics[width=\columnwidth]{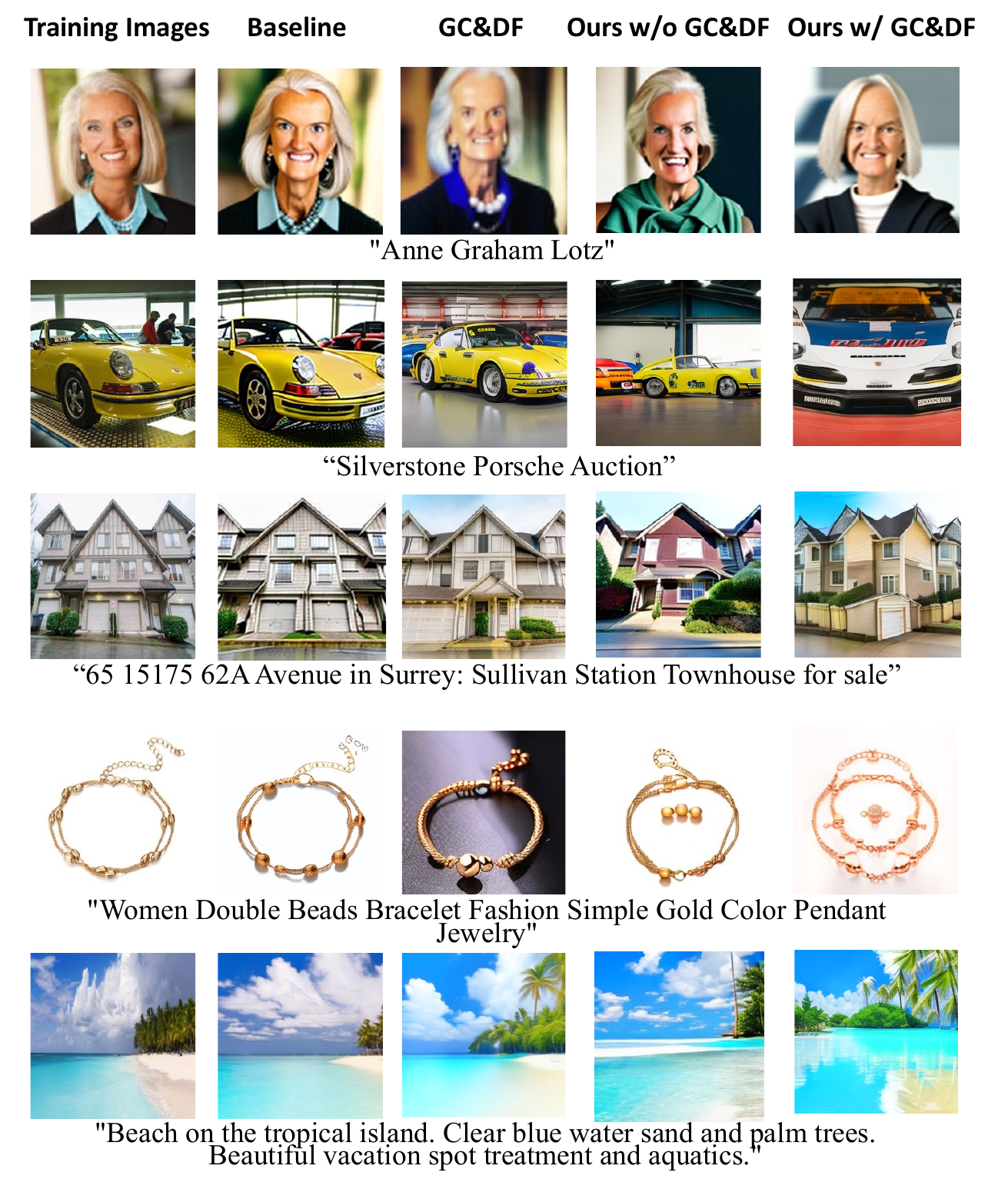}
      \caption{Images in the first row are from training dataset that are replicated. Images in the second row are generated using fine-tuned baseline model, %Stable Diffusion v2.1~\cite{rombach2022high}
      which show some replication with training images. Images in the third row are generated by our best performance model, with less content replicated with training images.}
      \label{fig:mitigating-examples}
\end{figure}

\begin{table*}[tbhp]
\centering
\resizebox{\textwidth}{!}{%
\begin{tabular}{c|c|c|ccccccccccccccc}
\hline\hline
       & $R_{real}$   & Baseline & \multicolumn{15}{c}{LoyalDiffusion + Generalized Captions + Dual-Fusion}                                                                                                                    \\ \hline
$w_{lat}$ & -     & -        & \multicolumn{5}{c|}{0.1}                                   & \multicolumn{5}{c|}{0.1}                                   & \multicolumn{5}{c}{0.1}               \\ \hline
$w_{emb}$ & -     & -        & \multicolumn{5}{c|}{0.1}                                   & \multicolumn{5}{c|}{0.3}                                   & \multicolumn{5}{c}{0.5}               \\ \hline
$\tau$      & -     & -        & 100   & 300   & 500   & 700   & \multicolumn{1}{c|}{900}   & 100   & 300   & 500   & 700   & \multicolumn{1}{c|}{900}   & 100   & 300   & 500   & 700   & 900   \\ \hline
R$\downarrow$      & 0.275 & 0.617    & 0.384 & 0.394 & 0.430 & 0.531 & \multicolumn{1}{c|}{0.562} & 0.341 & 0.351 & 0.363 & 0.443 & \multicolumn{1}{c|}{0.484} & 0.316 & \textbf{0.316} & 0.321 & 0.365 & 0.392 \\ \hline
FID$\downarrow$    & -     & 18.01    & 18.42 & 18.07 & 17.89 & 15.92 & \multicolumn{1}{c|}{14.79} & 21.03 & 18.15 & 18.42 & 17.22 & \multicolumn{1}{c|}{17.00} & 21.11 & \textbf{18.98} & 19.51 & 18.44 & 18.12 \\ \hline
CLIP$\uparrow$   & -     & 0.313        & 0.293     & 0.304     & 0.312     & 0.315     & \multicolumn{1}{c|}{0.316}     & 0.292     & 0.306     & 0.313     & 0.318     & \multicolumn{1}{c|}{0.318}     & 0.288     & \textbf{0.308}     & 0.313     & 0.318     & 0.319     \\ \hline \hline
\end{tabular}}
\caption{Results of layering GC\&DF \cite{li2024mitigate} with the proposed \textit{LoyalDiffusion} framework.}
\label{tab:result-with-GCDF}
\end{table*}

Building on previous work \cite{li2024mitigate}, we incorporate generalized captions (GC) and dual-fusion (DF) to further mitigate replication. For dual-fusion, we perform image fusion in the latent space using a weight $w_{lat} = 0.1$ and text fusion at the embedding level with $w_{emb} = 0.1, 0.3, 0.5$ \cite{li2024mitigate}. 
% We explore the combination of these techniques with two main configurations: (1) using only the de-replicate U-Net, and (2) employing multi-stage training with the de-replicate U-Net. 
The experimental results are presented in Table \ref{tab:result-with-GCDF}.

% \textbf{Analysis} 
The combination of GC\&DF with LoyalDiffusion significantly improves the performance. With $w_{lat}=0.1$, $w_{emb}=0.5$ and $\tau=300$, it achieves a reduction in replication of $48.78\%$ with only $0.97$ FID and a $0.005$ CLIP score loss, compared with the baseline. Furthermore, even with a strict FID that should be almost the same as the baseline, our method still achieves a $43.11\%$ reduction of replication, with $w_{lat}=0.1$, $w_{emb}=0.3$ and $\tau=300$. These results support the claim that GC\&DF and our architectural modifications complement each other, 
GC\&DF mitigates replication from the data perspective, while our methods target the model structure itself. %Together, they work to reduce replication without negatively impacting model performance.

\subsection{Visual Comparison}

We present a visual comparison to show the effectiveness of our approach in Figure~\ref{fig:mitigating-examples}, which includes original images from the training dataset, the most likely replicated image generated by the baseline model, and  our best-performing model. The comparison illustrates a significant reduction of replication.
% , demonstrating that our method effectively limits repetitive patterns in the generated outputs.

% \subsection{Analysis trade-off between R and FID}

\subsection{Ablation Studies}
% Need to just conduct an ablation study by selectively removing or modifying each part and assessing its impact on model performance? Or need more detailed analysis?

% \textbf{Use RAU-net for all timesteps}

\begin{table}[]
\centering
\resizebox{\columnwidth}{!}{%
\begin{tabular}{c|ccc|cc}
\hline\hline
\multirow{2}{*}{Model}   & \multicolumn{3}{c|}{Ablation}                                                    & \multicolumn{2}{c}{Metrics} \\ \cline{2-6} 
                         & RAU-Net & \begin{tabular}[c]{@{}c@{}}Two-stage\\ training\end{tabular} & GC\&DF & R$\downarrow$            & FID$\downarrow$          \\ \hline
Baseline                 & \XSolidBrush       & \XSolidBrush                                                             & \XSolidBrush      & 0.617        & 18.01        \\ \hline
\multirow{5}{*}{Ablated} & \Checkmark       & \XSolidBrush                                                             & \XSolidBrush      & 0.386        & 21.31        \\ \cline{2-6} 
                         & \XSolidBrush       & \Checkmark                                                             & \XSolidBrush      & N/A          & N/A          \\ \cline{2-6} 
                         & \XSolidBrush       & \XSolidBrush                                                             & \Checkmark      & 0.419        & 17.16        \\ \cline{2-6} 
                         & \Checkmark       & \Checkmark                                                             & \XSolidBrush      & 0.378        & 19.17        \\ \cline{2-6} 
                         & \Checkmark       & \XSolidBrush                                                             & \Checkmark      & 0.313        & 21.39        \\ \hline
Ours                     & \Checkmark       & \Checkmark                                                             & \Checkmark      & 0.316        & 18.98        \\ \hline\hline
\end{tabular}%
}
\caption{Results of ablation studies.
% (\XSolidBrush indicates the module is suppressed and \Checkmark means the module is applied)
}
\label{tab:ablation-1}
\vspace{-3mm}
\end{table}

Table \ref{tab:ablation-1} shows the ablation studies. Applying RAU-Net among all timesteps reduces replication comparing with baseline but increases FID, indicating lower fidelity. When two-stage training strategy is introduced, it significantly recovers FID without impacting replication reduction. Additionally, LoyalDiffusion achieve a better replication score than GC\&DF and combining LoyalDiffusion with GC\&DF outperforms only using one of them.

\textbf{FID-Centric vs. R-Centric Two-Stage Training.}
In the previous section, we showed that early timesteps (when $t$ is large) have less impact on FID which motivated applying RAU-Net to those timesteps. We refer to this approach as FID-centric. However, the similarity curve in Figure \ref{fig:two-stage-approach} can also serve as a measure of replication, which shows that similarity increases fast during late timesteps (when $t$ is small). So later timesteps are critical to generate replicated images which could to some degree motivate applying RAU-Net to later timesteps, an approach we refer to as R-centric. For this reason, we conduct an ablation experiment in which we apply $U_{ra}$ when $t\leq\tau$ and $U_{st}$ when $t>\tau$. The results are shown in Table \ref{tab:two-stage-result_2} and confirm that the FID-centric method performs better than the R-centric method in both replication reduction and image fidelity.

% * Ideal: Start with RA-Unet for high timesteps use Unet for low timesteps. Compare with UNET for hightime steps and RA-Unet for low time steps. [SKIP]

% \textbf{With and w/o Dual Fusion}

% * Train both RA-Unet and Baseline Unet for all timesteps even though they are used on a subset of timesteps... []

% * Use two baselines trained with different ranges [SKIP]

\begin{table}
\centering
\resizebox{0.95\columnwidth}{!}{%
\begin{tabular}{c|ccccc}
\hline\hline
$\tau$ & 100 & 300    & 500 & 700    & 900 \\ \hline
R$\downarrow$                & 0.610   & 0.576 & 0.511   & 0.445 & 0.384   \\ \hline
FID$\downarrow$              & 18.93   & 19.27 & 23.11   & 21.58 & 20.88   \\ \hline
CLIP$\uparrow$             & 0.311   & 0.301      & 0.285   & 0.282      & 0.288   \\ \hline \hline
\end{tabular}%
}
\caption{Results for replication-centric two-stage training, with $U_{ra}$ for $t\leq \tau$ and $U_{st}$ for $t>\tau$.}
\label{tab:two-stage-result_2}
\vspace{-2mm}
\end{table}

% \subsection{Discussion}
\section{Conclusions and Future Work}
% Give insight of skip connection, maybe have other structure.
In this work, we introduced \textit{LoyalDiffusion}, a novel framework designed to mitigate replication risks in diffusion models while preserving image generation quality. Our approach leverages a replication-aware U-Net (RAU-Net) that selectively incorporates convolutional information transfer blocks at specific skip connections, effectively balancing quality and privacy. Additionally, we propose a novel training strategy applies RAU-Net selectively across timesteps, allowing for targeted mitigation of replication during less quality-sensitive stages of the generation process. Through extensive experiments, \textit{LoyalDiffusion} demonstrated a significant reduction in replication scores compared to baseline models and prior state-of-the-art methods, achieving this with minimal impact on FID and generation quality. 

% By addressing replication at the model structure level, \textit{LoyalDiffusion} provides a foundational advancement in privacy-preserving generative models, marking a meaningful step toward responsible AI-driven content generation. 

While our approach with RAU-Net demonstrates promising results, there are limitations to our current design. The RAU-Net architecture, specifically the choice and placement of convolutional information transfer blocks, was determined through empirical evaluation of a limited set of transfer block configurations. Consequently, it is possible that other information transfer mechanisms could yield better results.
%in reducing replication while maintaining high image quality. 
Future research could explore a broader range of information transfer blocks or optimize RAU-Net architecture through automated search techniques to achieve an optimal balance between replication reduction and fidelity. Despite these limitations, our work represents a critical advancement for the field. In particular, to the best of our knowledge, this study is the first to explicitly examine the link between skip connections in diffusion models and training data leakage, thereby providing a new perspective on mitigating privacy risks within generative models.

%\clearpage
{
    \small
    \bibliographystyle{ieeenat_fullname}
    \bibliography{main}
}

% WARNING: do not forget to delete the supplementary pages from your submission 
% \clearpage
% \setcounter{page}{1}
% \maketitlesupplementary

% \section{Rationale}
% \label{sec:rationale}
% % 
% Having the supplementary compiled together with the main paper means that:
% % 
% \begin{itemize}
% \item The supplementary can back-reference sections of the main paper, for example, we can refer to \cref{sec:intro};
% \item The main paper can forward reference sub-sections within the supplementary explicitly (e.g. referring to a particular experiment); 
% \item When submitted to arXiv, the supplementary will already included at the end of the paper.
% \end{itemize}
% % 
% To split the supplementary pages from the main paper, you can use \href{https://support.apple.com/en-ca/guide/preview/prvw11793/mac#:~:text=Delete%20a%20page%20from%20a,or%20choose%20Edit%20%3E%20Delete).}{Preview (on macOS)}, \href{https://www.adobe.com/acrobat/how-to/delete-pages-from-pdf.html#:~:text=Choose%20%E2%80%9CTools%E2%80%9D%20%3E%20%E2%80%9COrganize,or%20pages%20from%20the%20file.}{Adobe Acrobat} (on all OSs), as well as \href{https://superuser.com/questions/517986/is-it-possible-to-delete-some-pages-of-a-pdf-document}{command line tools}.

\clearpage
\setcounter{page}{1} % Reset page numbering
\renewcommand{\thesection}{\Alph{section}} % Change section numbering to letters
\setcounter{section}{0} % Reset section counter to start at A
\maketitlesupplementary

\section{Appendix}
\label{sec:Appendix}

\subsection{Additional Experimental Results}
\label{sec:Appendix-1}

In this section, we provide additional experimental results focusing on different combinations of modified skip connections (SC) with information transfer blocks, extending the analysis presented in Section~\ref{sec:3-1}.  Section~\ref{sec:3-1} analyzed a limited combinations of applying different information transfer blocks to different skip connections (SC) and determined applying \textit{Conv} to SC $3$ and SC $4$ simultaneously gives the best replication reduction without increasing FID. Here, we explore additional combinations of modified skip connections, as shown in Table~\ref{tab:various-SC-result}.

\begin{table*}[]
\centering
\resizebox{0.95\textwidth}{!}{
\begin{tabular}{l|ccccccccccr}
\hline  \hline
    \multicolumn{11}{c}{Removing Skip Connection}     \\ \hline
% RSK             &       &       &       &       &       &       \\
SC \#      & 1\&2     & 1\&3     & 1\&4     & 2\&3     & 2\&4   &   3\&4    &     1\&2\&3    &    1\&2\&4    &    1\&3\&4    &    2\&3\&4    &    \\\hline
R$\downarrow$               & 0.247 & 0.547 & 0.562 & 0.499 & 0.585 & 0.604 & 0.226 & 0.243 & 0.543 & 0.519 &       \\\hline
FID$\downarrow$            & 105.76 & 24.21 & 22.09 & 33.62 & 22.70 & 20.30 & 141.38 & 102.80 & 25.67 & 35.31 &       \\ \hline \hline
 \multicolumn{11}{c}{\textit{MP}}   \\\hline
SC \#      & 1\&2     & 1\&3     & 1\&4     & 2\&3     & 2\&4   &   3\&4    &     1\&2\&3    &    1\&2\&4    &    1\&3\&4    &    2\&3\&4    &    \\\hline
R$\downarrow$               & 0.600 & 0.609 & 0.613 & 0.596 & 0.606 & 0.614 & 0.597 & 0.611 & 0.590 & 0.602 &       \\\hline
FID$\downarrow$            & 19.20 & 16.85 & 18.05 & 18.11 & 18.50 & 16.52 & 18.68 & 18.79 & 17.72 & 17.40 &       \\ \hline \hline
 \multicolumn{11}{c}{\textit{Conv}}  \\ \hline
SC \#      & 1\&2     & 1\&3     & 1\&4     & 2\&3     & 2\&4   &   3\&4    &     1\&2\&3    &    1\&2\&4    &    1\&3\&4    &    2\&3\&4    &    \\\hline
R$\downarrow$               & 0.311 & 0.303 & 0.458 & 0.345 & 0.566 & \textbf{0.386} & 0.235 & 0.326 & 0.272 & 0.308 &       \\\hline
FID$\downarrow$            & 89.70 & 50.85 & 44.47 & 41.90 & 22.00 & \textbf{21.31} & 142.01 & 78.51 & 52.88 & 50.56 &       \\ \hline \hline
 \multicolumn{11}{c}{\textit{Multi-Conv}}      \\ \hline
SC \#      & 1\&2     & 1\&3     & 1\&4     & 2\&3     & 2\&4   &   3\&4    &     1\&2\&3    &    1\&2\&4    &    1\&3\&4    &    2\&3\&4    &    \\\hline
R$\downarrow$               & 0.368 & 0.455 & 0.541 & 0.412 & 0.579 & 0.460 & 0.243 & 0.369 & 0.399 & 0.386 &       \\\hline
FID$\downarrow$            & 68.25 & 26.39 & 25.45 & 34.24 & 20.28 & 20.74 & 116.27 & 63.14 & 28.04 & 42.21 &       \\ \hline \hline
\end{tabular}}
\caption{Replicaiton score and FID for different information transfer blocks on LAION-10K~\cite{schuhmann2022laion}. The 'SC \#' indicates the information transfer block is only applied to the specific combination of skip connections numbered as in Figure~\ref{fig:overview}(a)}
\label{tab:various-SC-result}
\end{table*}

The results reveal two distinct trends: in some cases, replication is significantly reduced, but at the cost of a notable degradation in FID; in other cases, the impact on FID is minimal or even slightly improved, but the reduction in replication is negligible. Only two configurations achieved satisfactory results: applying the \textit{Conv} information transfer block to SC $3$ and $4$, and applying the \textit{Multi-Conv} information transfer block to SC $3$ and $4$. Both configurations effectively reduced replication while maintaining almost unchanged FID. Among these, the \textit{Conv} information transfer block applied to SC $3$ and $4$ demonstrated a more pronounced reduction in replication than \textit{Multi-Conv}, validating the choice made in Section~\ref{sec:3-1}.
%to adopt the \textit{Conv} information transfer block on SC $3$ and $4$ as the primary structure of RAU-Net.

\subsection{Visualize Generated Images for Various FIDs}
\label{sec:Appendix-2}

\begin{figure*}[t]
      \centering
      \includegraphics[width=\textwidth]{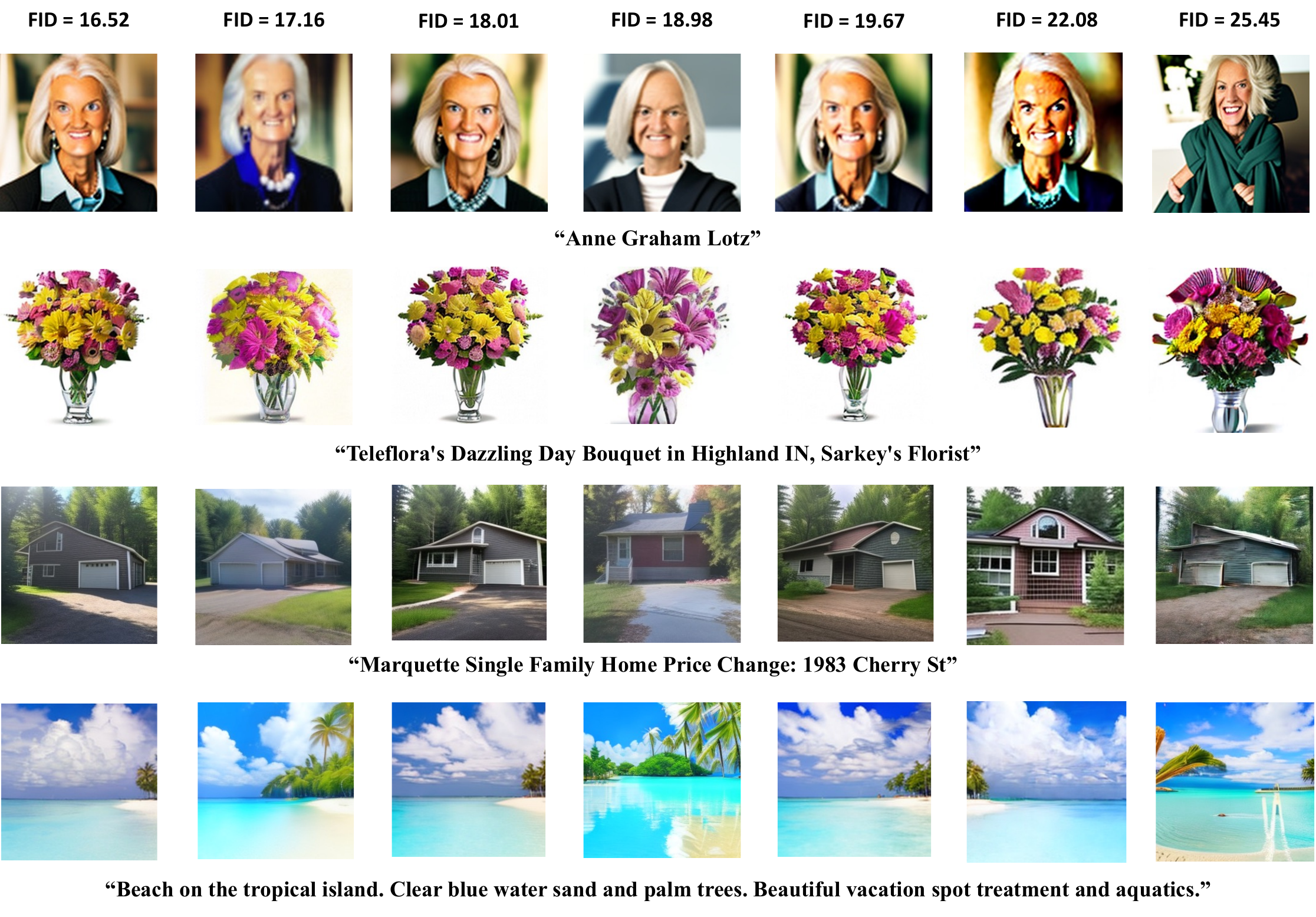}
      \caption{Examples of generated images with different FIDs. The images come from the following models: 
      column (1) MP 3\&4 from Table \ref{tab:various-SC-result}; column (2) GC\&DF from Table \ref{tab:compare-prior}; 
      column (3) Baseline from Table \ref{tab:compare-prior}; 
      column (4) $w_{lat} = 0.1, w_{emb}=0.5,$ and $\tau = 300$ from Table \ref{tab:result-with-GCDF}; 
      column (5) Remove skip connection 4 from Table \ref{tab:raunet-result-2}; 
      column (6) Two-stage result with $\tau = 100$ from Table \ref{tab:two-stage-result_1}; 
      and (7) Multi-Conv 1\&4 from Table \ref{tab:various-SC-result}.}
      \label{fig:fid-figure}
\end{figure*}

In this section, we provide visualizations of generated images corresponding to different FID values, as shown in Figure~\ref{fig:fid-figure}. The third column shows images generated by baseline model with FID$=18.01$. The visual examples illustrate that when the FID is within the range of ${18.01} \pm 2$, the generative ability remains comparable, producing visually similar and acceptable images. However, with FID larger than $20$, the image quality drops significantly. This demonstrates that an FID within $(16, 20)$ does not significantly impact image quality, making it an acceptable range for evaluating the performance of our proposed \textit{LoyalDiffusion}.

\subsection{\textbf{\textit{RepliBing}} Dataset}
\label{sec:Appendix-3}

To further evaluate replication in generative models, we created the \textit{RepliBing} dataset mentioned in Section \ref{sec:RepliBing}. This dataset is constructed using the Bing search engine to retrieve publicly available images from the internet, based on the same prompts used during training and inference. \textit{RepliBing} is designed to provide a diverse set of images that align closely with the input captions while minimizing the chance of direct replication of the training data.

\begin{figure*}[t]
      \centering
      \includegraphics[width=\textwidth]{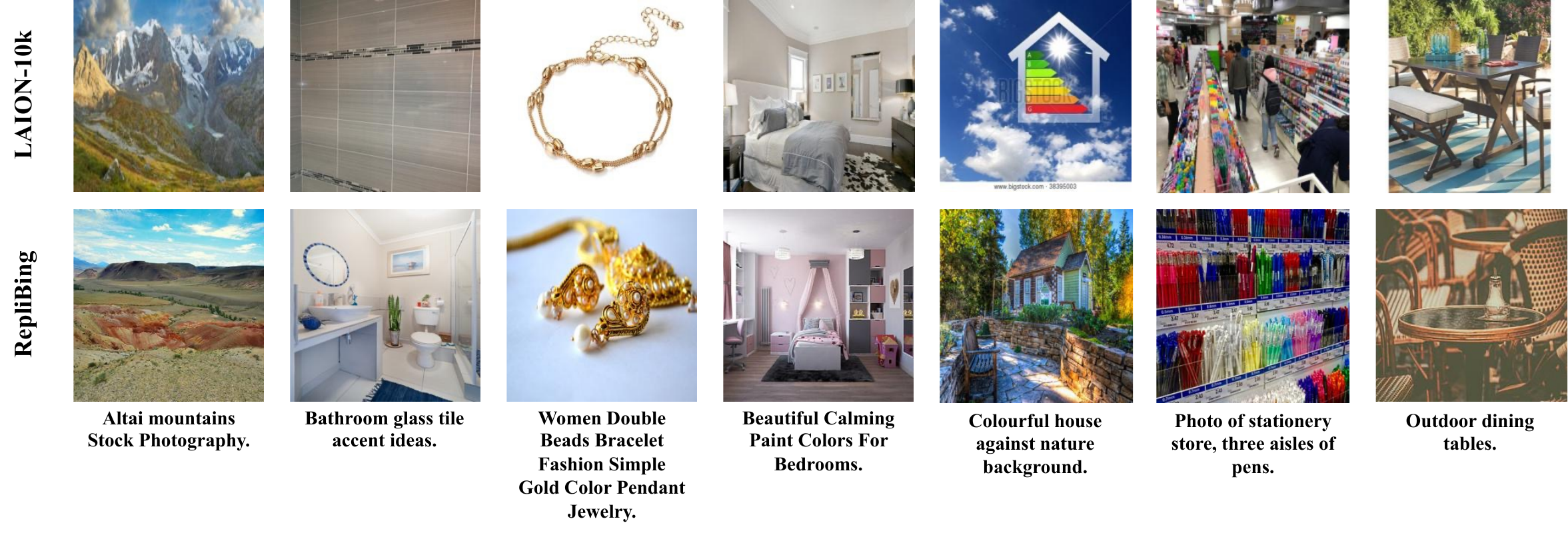}
      \caption{Examples of images from \textit{RepliBing} sharing the same caption with images from LAION-10K.}
      \label{fig:replibing}
\end{figure*}

Figure~\ref{fig:replibing} showcases examples from the \textit{RepliBing} dataset alongside corresponding training images that share the same captions. As shown, the \textit{RepliBing} dataset captures a broad spectrum of visual diversity while maintaining fidelity to the provided prompts. These comparisons highlight the visual similarity in the conceptual alignment between \textit{RepliBing} images and the training data, without any direct replication. We plan to open-source this dataset to facilitate further research in mitigating replication in generative models upon acceptance of the paper. 
%This dataset will provide the community with a valuable resource for evaluating and mitigating replication in generative models.

\end{document}